# Flexible image analysis for law enforcement agencies with deep neural networks to determine: where, who and what

Henri Bouma [1*], Bart Joosten [1], Maarten Kruithof [1], Maaike de Boer [1],
Alexandru Ginsca [2], Benjamin Labbe [2], Quoc T. Vuong [2]

[1] TNO, The Hague, The Netherlands.
[2] CEA LIST, Gif-sur-Yvette, France.

## ABSTRACT

Due to the increasing need for effective security measures and the integration of cameras in commercial products, a huge amount of visual data is created today. Law enforcement agencies (LEAs) are inspecting images and videos to find radicalization, propaganda for terrorist organizations and illegal products on darknet markets. This is time consuming. Instead of an undirected search, LEAs would like to adapt to new crimes and threats, and focus only on data from specific locations, persons or objects, which requires flexible interpretation of image content. Visual concept detection with deep convolutional neural networks (CNNs) is a crucial component to understand the image content. This paper has five contributions. The first contribution allows image-based geo-localization to estimate the origin of an image. CNNs and geotagged images are used to create a model that determines the location of an image by its pixel values. The second contribution enables analysis of fine-grained concepts to distinguish sub-categories in a generic concept. The proposed method encompasses data acquisition and cleaning and concept hierarchies. The third contribution is the recognition of person attributes (e.g., glasses or moustache) to enable query by textual description for a person. The person-attribute problem is treated as a specific sub-task of concept classification. The fourth contribution is an intuitive image annotation tool based on active learning. Active learning allows users to define novel concepts flexibly and train CNNs with minimal annotation effort. The fifth contribution increases the flexibility for LEAs in the query definition by using query expansion. Query expansion maps user queries to known and detectable concepts. Therefore, no prior knowledge of the detectable concepts is required for the users. The methods are validated on data with varying locations (popular and non-touristic locations), varying person attributes (CelebA dataset), and varying number of annotations.

**Keywords:** Deep learning, concept detection, convolutional neural networks, image-based geo-localization, fine-grained concept detection, person attributes, active learning, query expansion.

## 1. INTRODUCTION

Due to the increasing need for effective security measures and the integration of cameras in commercial products, a huge amount of visual data is created today. Law enforcement agencies (LEAs) are inspecting images and videos to find radicalization, propaganda for terrorist organizations and illegal products on darknet markets. This is time consuming. Instead of an undirected search, LEAs would like to adapt to new crimes and threats, and focus only on data from specific locations, persons or objects, which requires flexible interpretation of image content. Visual concept detection with deep convolutional neural networks (CNNs) is a crucial component to understand the image content.

This paper has five contributions. The first contribution allows image-based GPS localization to estimate the origin of an image. CNNs and geotagged images are used to create a model that determines the location of an image by its pixel values. The second contribution enables analysis of fine-grained concepts to distinguish sub-categories in a generic concept. The proposed method encompasses data acquisition and cleaning and concept hierarchies. The third contribution is the recognition of person attributes (e.g., glasses or moustache) to enable query by textual description for a person. The person-attribute problem is treated as a specific sub-task of concept classification. The fourth contribution is an intuitive image annotation tool based on active learning. Active learning allows users to define novel concepts flexibly and train CNNs with minimal annotation effort. The fifth contribution increases the flexibility for LEAs in the query definition by using query expansion. Query expansion maps user queries to known and detectable concepts. Therefore, no prior knowledge of the detectable concepts is required for the users.

[*] henri.bouma@tno.nl; phone +31 888 66 4054; http://www.tno.nl

The outline of this paper is as follows. Sections 2, 3 and 4 answer the 'Where', 'What' and 'Who' questions respectively for LEAs. Section 2 describes the image-based geo-localization ('Where was this image taken?'). Section 3 describes fine-grained detection ('What is visible in the image?'). Section 4 shows how person attributes can be computed ('Who is present at this image?'). Sections 5 and 6 aim for higher flexibility for the LEAs. Section 5 describes the interactive annotation tool, which allows users to train new concepts (Section 5). And Section 6 shows query expansion to allow search-by-keywords, even if the keywords are not identical to the trained concepts.

## 2. IMAGE-BASED GEO-LOCALIZATION

Image geo-localization is often addressed as an image retrieval problem and GPS is inferred from top retrieval results. Early systems rely on local features such as SIFT [26] to select the closest images from a database. From the top-matched candidates, Zhang [37] performs camera motion estimation to determine exact location based on the best match reference image. On the other hand, IM2GPS [16] extracts global descriptors as features from images and stores them in a large database. They also evaluate the geolocation accuracy when using each of the selected features in isolation and in unison to tell which of them are discriminative for this task. With the recent rise of CNNs for computer vision problems, many works have managed to employ them specifically for geo-localization task in different manners. Most of them start from pre-trained networks that are efficient for object classification. Arandjelovic [1] proposes a novel pooling layer, called NetVLAD, built on top of last convolutional layers of well-known CNN architectures. This is the closest work to ours as they also use Google Street View imagery and triplet loss for training their model. Lin [23] and Vo [36] also train end-to-end CNN for place image retrieval. However, their works aim at cross-view images, matching ground level with aerial view images. Kim [19] and Noh [28] focus on an attention mechanism, adding novel layers on top of existing networks to assign different weights on image regions according to their importance in place discrimination. We look at the image geo-localization problem as an image retrieval task but, departing from existing works, we strive for open world geo-localization and propose an end-to-end pipeline in which new locations (e.g. cities, countries) can be easily introduced.

**2.1 Method**

Our image dataset for GPS localization is collected from Google Streetview[*]. This guarantees large coverage in many regions world-wide and uniform distribution of images with reasonable quality. In details, this dataset covers 32 European cities where Google Streetview is available.

First, we need to prepare GPS coordinates as required parameters of Streetview image requests. For each city, we download OpenStreetMap (OSM) data corresponding to the smallest region enclosing the selected area. Then, OSM objects concerning the city are filtered using its polygon boundary. Nodes with GPS coordinates representing streets are extracted using filters with OSM highway tag. It is worth noting that there is a huge difference in the number of GPS points resulting from this step between areas, depending on their surface area as well as their popularity.

Second, the same number of coordinates are sampled from the extracted list, aiming for similar size for each image collection. To promote well-spread geographical distribution of sample points, we perform K-Means clustering on GPS coordinates of each area to find $K$ clusters as well as their centroids, with $K$ being the targeted number of sample points for the area. Then, we sample a single data point for each cluster, using it for downloading images of the corresponding location from Google Streetview. As shown in Figure 1, sampling coordinates from clusters yields more even geographical distribution than random sampling by avoiding cluttering in popular areas with many OSM nodes.

Third, with sampled GPS points, we make direct requests to the Google Streetview API to download their images. In detail, for each point, we attempt to collect images of 4 view directions from the same location. Other parameters include 120° for field-of-view and 0° for camera pitch. Given these parameters, the image result of each request is deterministic. The geo-localization of an image can be addressed as an image retrieval problem and GPS inference is based on top results. Gordo [15] proposed an end-to-end approach for image retrieval using CNNs. Image embeddings extracted from this architecture, called Deep Image Retrieval (DIR) descriptors are robust for object matching, especially landmarks, since the network was trained on a landmarks dataset [2]. We propose another method to represent images for retrieval. First, we fine-tune a pre-trained ImageNet model for the classification of our Streetview dataset (each city corresponds to a class) so that it adapts to images of streets and landmarks. Then, we could employ its feature layer to extract fixed-size representations of our query and index images. For any choice of image embedding space, we perform k-NN search with product quantization [18] for index images that are nearest to the query one. We then infer the query image GPS using the average of GPS of top results amongst index images.

---

[*] https://www.google.com/streetview/

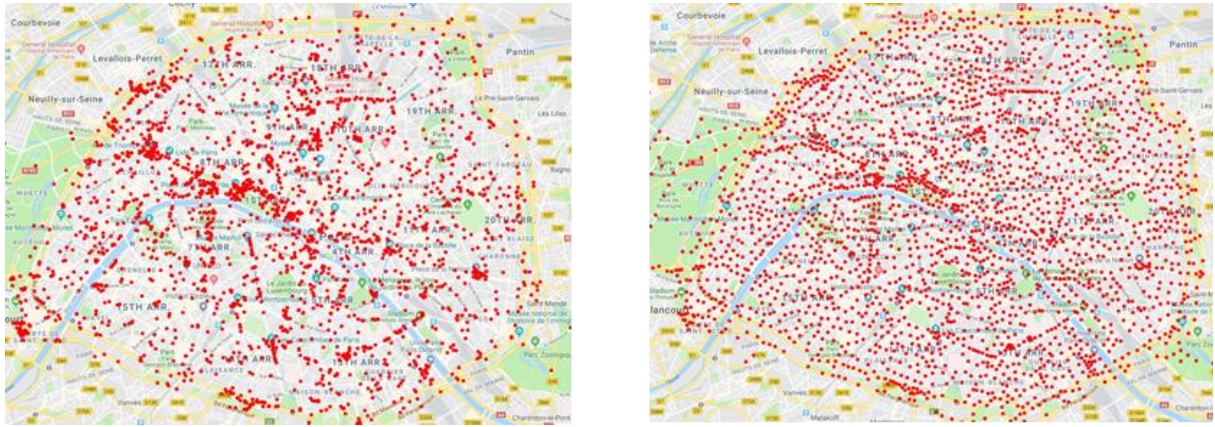

Figure 1. Different sampling strategies illustrated for Paris. Without clustering (left), with clustering (right).

## 2.2 Experiments and results

Our dataset includes 32 cities, each with about 10000 images, partitioned into training and testing set with a 8:2 ratio. To ensure no data leakage, the partitioning is geography-based, that is, for each city, we divide the GPS space into a grid and assign all images of the same cell to either training or testing set (see Figure 2). The query and index images are drawn from the testing set in random manner.

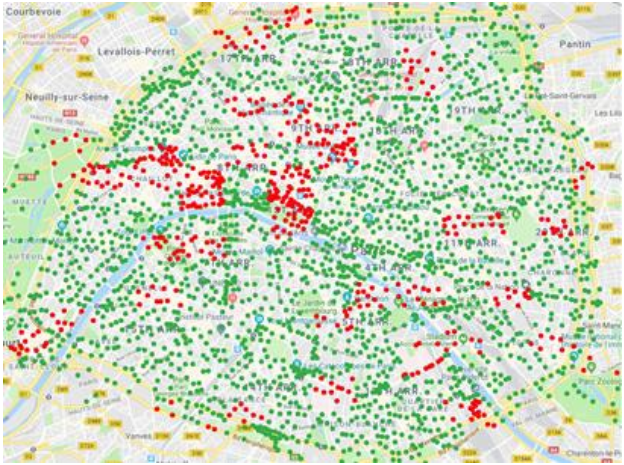

Figure 2. Selection of training (green points) and validation samples (red points) illustrated for Paris.

For our proposed method, we fine-tuned a pre-trained ResNet101 [17] for classification on 32 cities until convergence. The penultimate layer is employed for feature extraction. Table 1 shows our evaluation for both methods on our Streetview testing dataset. We report average distance error (in km) from the ground truth, geo-localization accuracy within 1 km, 25 km, 200 km respectively, for different choices of number of nearest neighbors. Since our dataset consists of images from cities of different countries, mis-prediction of a city would heavily penalize these scores. We notice from these scores that in most cases, the first match is already close to the ground-truth, as shown by degradation in terms of geo-localization accuracy at different scales when accounting for additional matches. Some sample queries and their top matches found by our proposed method are illustrated in Figure 3.

Table 1: Results for DIR and ResNet101-32 descriptors.

| Descriptor | NN | Dist. Error | Geo. Acc. | | |
|---|---|---|---|---|---|
| | | | 1 km | 25 km | 200 km |
| DIR | 1 | 779.25 | 0.105 | 0.302 | 0.332 |
| DIR | 5 | 672.24 | 0.013 | 0.052 | 0.163 |
| DIR | 9 | 661.77 | 0.006 | 0.028 | 0.146 |
| ResNet101-32 | 1 | 214.06 | 0.234 | 0.764 | 0.791 |
| ResNet101-32 | 5 | 209.19 | 0.087 | 0.544 | 0.702 |
| ResNet101-32 | 9 | 213.31 | 0.063 | 0.463 | 0.693 |

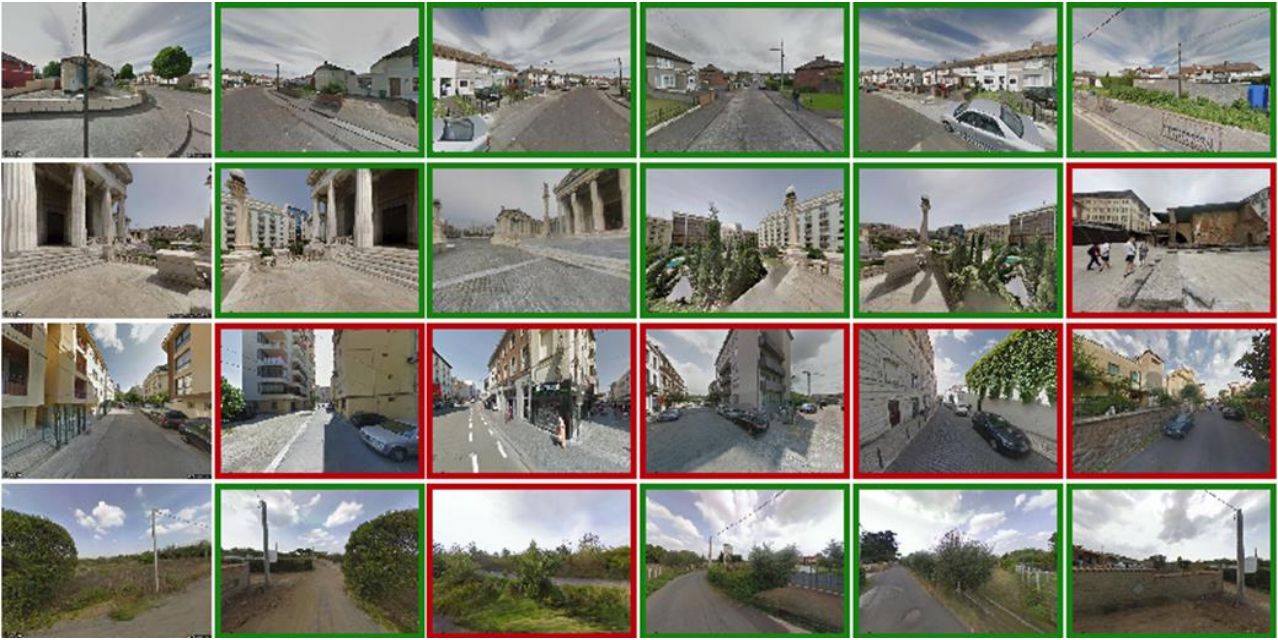

Figure 3. Example of retrieval results.

### 2.3 Conclusions

In this section, we presented an end-to-end pipeline for open world image geo-localization through the use of Street View images. We showed that it is possible to automatically gather an image dataset starting from a list of target locations, for which we proposed a novel clustering based GPS coordinates sampling method. This process led to the construction of a new geo-localization dataset that covers 32 European capitals. In order to infer the coordinates of a novel image, we illustrated the effectiveness of an image retrieval approach, which was implemented with both a state of the art image descriptor and a novel image representation that was specifically designed for Street View images.

### 3. FINE-GRAINED DETECTION

CNN training is usually done with manually labeled data but this approach has the obvious disadvantage that large volumes of images need to be validated by humans. In an attempt to overcome this problem, focus is put on the possibility to exploit large Web image corpora instead of manually labelled datasets. Prior work [33], done outside the deep-learning field proposes re-ranking techniques that rely on cross-validation scores of web images. A challenge related to this unsupervised re-ranking process is that it only works if noisy images are not predominant in the initial dataset. Image classification is predominantly performed using "bottom-up" CNN features in which concept representations are progressively abstracted

4 / 16

from the raw content of the images. The term 'concept' is an established term in the multimedia analysis and computer vision research communities and is typically associated with a topic, entity, object or theme depicted in an image. Semantic features [3] encode images as a series of visual concepts that are fed into the pipeline. Here, we rely on the pipeline used for building semantic features to train individual binary classifiers for fine-grained recognitions of firearms and armored vehicles.

### 3.1 Method

We follow the training protocol of the Semfeat descriptor [14], a scalable semantic image descriptor built on top of midlevel CNN features. These were learned using a training dataset that was obtained through a manual annotation process that is difficult to scale up. As an alternative, we propose here a full learning pipeline that exploits Web images for CNN training instead of a manually built dataset. We start from a manually curated list of 314 firearms and 313 armored vehicles. For each concept, images are automatically gathered using the Bing image search API[*]. Then, we remove corrupt images. In Figure 4, we show the distribution of the number of available training images for both collections. We can observe that for both datasets, we have few training samples for most of the concepts. This is understandable, as we searched for very specific firearm or vehicles models that may have few different visual depictions (e.g. *Remington Model Seven LS, Mossberg MVP Light Chassis Rifle*).

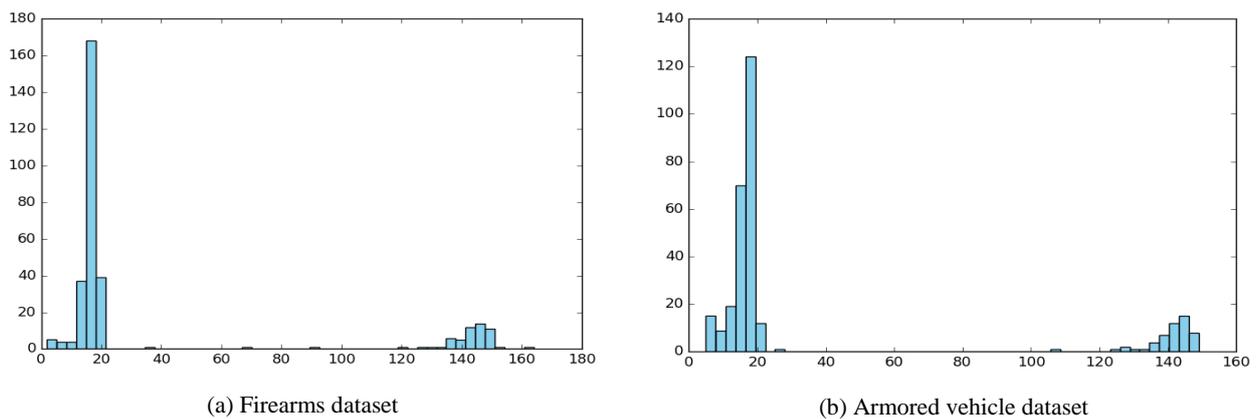

(a) Firearms dataset                (b) Armored vehicle dataset

Figure 4. Distributions of the number of available training images per concept.

Our main objective is to propose a fine grained concept detection approach that is applicable to the very large number of concepts which can appear in the firearms and armored vehicles domains. In order to achieve this objective, we adopt a feature transfer approach similar in spirit with that of [30] and [14], which ensures a good balance between concept detection accuracy and scalability. Feature transfer involves learning a deep learning model using a large set of initial concepts of general nature, e.g. "rifle", "pistol", "AK-47", etc. (typically in the range of 1,000) with a convolutional neural network and then exploiting the outputs of an intermediary layer of this model as features to model concepts that do not appear in the training set. The underlying assumption here is that the CNN model is able to generalize beyond its initial scope due to the fact that the initial concepts and the newly modeled ones share enough visual properties for a proper modeling of the latter. For instance, assuming that concepts "automatic rifle" and "AK-47" are contained in the set of initial concepts, then a new concept "AR-16" could be thought of as a combination of those concepts (and some additional ones). Detection accuracy is ensured through the use of efficient CNN models such as VGG.

Scalability is a second central requirement and it is dealt with in different forms:

- Scalable and efficient machine learning. Schroff [33] showed that techniques from the linear SVM family give good results in image classification based on local features derived from SIFTs. In initial experiments, we tested different linear solvers with CNN features and our results confirmed their conclusions. Confirming the robustness of the features no notable differences were observed between the solvers tested. Furthermore, we also tested non-linear SVMs (notably the RBF kernel) and a slight improvement is obtained but with a high computational cost for both training and testing phases. Consequently, a L2- regularized L2-loss SVM was consistently used in our experiments. This solver is particularly interesting at test time since the classification score of a concept is done through a simple dot product computation.

---
[*] https://azure.microsoft.com/en-us/services/cognitive-services/bing-image-search-api/

- Training with Web data. The public availability of ImageNet [13], a manually curated visual resource that includes over 14 million images illustrating 22,000 concepts, had a positive influence on research dealing with large scale concept detection. However, very few concepts are related to our domains of interest. We exploit this dataset only through the use of pre-trained models that we fine-tune on the Web images collections that we gathered as an even more comprehensive collection of visual data. This exploration is motivated by the fact that, while large, ImageNet does not cover concepts which are very relevant when dealing fine-grained topics.

- Concepts are learned independently from each other. In a majority of classification tasks, such as the ImageNet Challenge [32] it is assumed that all concepts are known in advance and a one-versus-all approach is used for learning. However, in practice it is often the case that new concepts need to be added to cope with user needs and, when working with tens of thousands of classes, relearning all models with a one-versus-all approach is highly impractical. Consequently, learning is done independently using a one-versus-many approach that is inspired by existing work from Bergamo [3] and Schroff [33].

### 3.2 Experiments and results

Using the pipeline detailed in the previous section, we train a total of 314 binary classifiers for fine-grained firearms concepts and 313 for armored vehicles. Given the imbalanced nature of the datasets and the small number of available training images (as shown in Figure 5), for each model we use all of the available positive images and a varying number of negatives selected among the rest of the classes. We depart from a classic one-versus-rest training configuration, as for a large portion of classes with few positive images, the large number of negatives hinders the learning process and pushes forward a large number of false negatives. We evaluated several configurations in which we chose the number of negatives as $k * positives$. We tested with $k$ in $\{2, 3, 5, 10, max\}$, where $max$ is equivalent with a one-versus-rest configuration. We use 5-fold cross validation and we look at the F1 score to set $k$. We found that $k=5$ gives the best balance between false positives and false negatives.

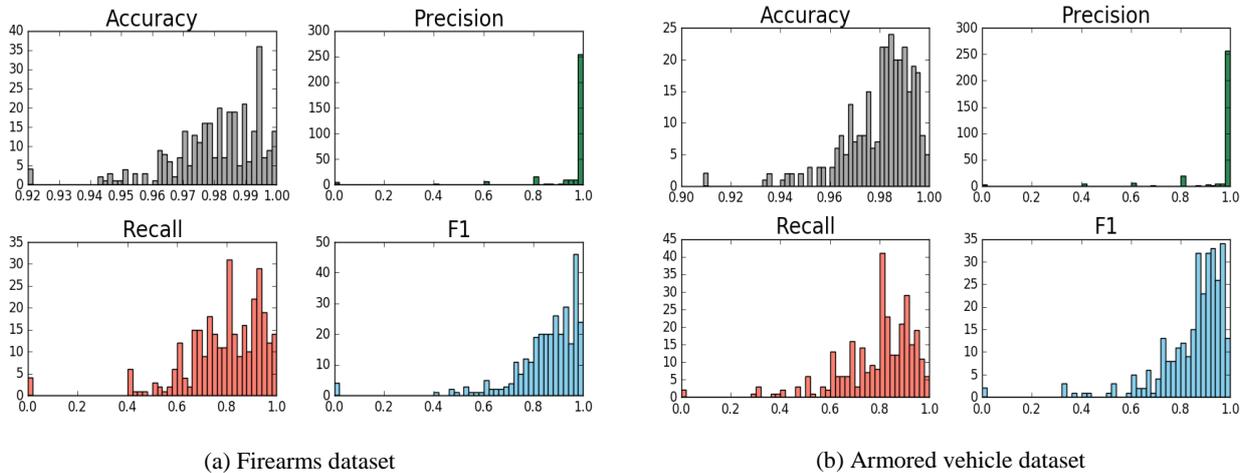

(a) Firearms dataset            (b) Armored vehicle dataset

Figure 5. Accuracy, precision, recall and F1 scores distributions for 314 firearms and 313 armored vehicles concept detectors.

In Figure 5, we provide a summary of accuracy, precision, recall and F1 scores distribution among all the concepts for the firearms and armored vehicles datasets. We observe that the precisions for both datasets are high, indicating that there are very few false positives. However, the recall is lower, which can be explained by the large number of false negatives induced by the high visual similarity among fine-grained concepts and the small number of positive training samples for the majority of the classes. When comparing datasets, we notice higher F1 scores for armored vehicles. Upon visual inspection, we can assume that this difference stems from the higher visual variability among armored vehicles compared to firearms.

A second evaluation is carried out using a 80-20 train-test split of training data and looking at the confusion matrix among concepts. In Figure 6, we chose to show a sample of the whole confusion matrix depicting 10 concepts for firearms and armored vehicles each, as the full confusion matrix would be less informative. We observed that for the firearms datasets, the *Rifle* concept tends to provide the highest confidence scores and to appear most often among the predictions.

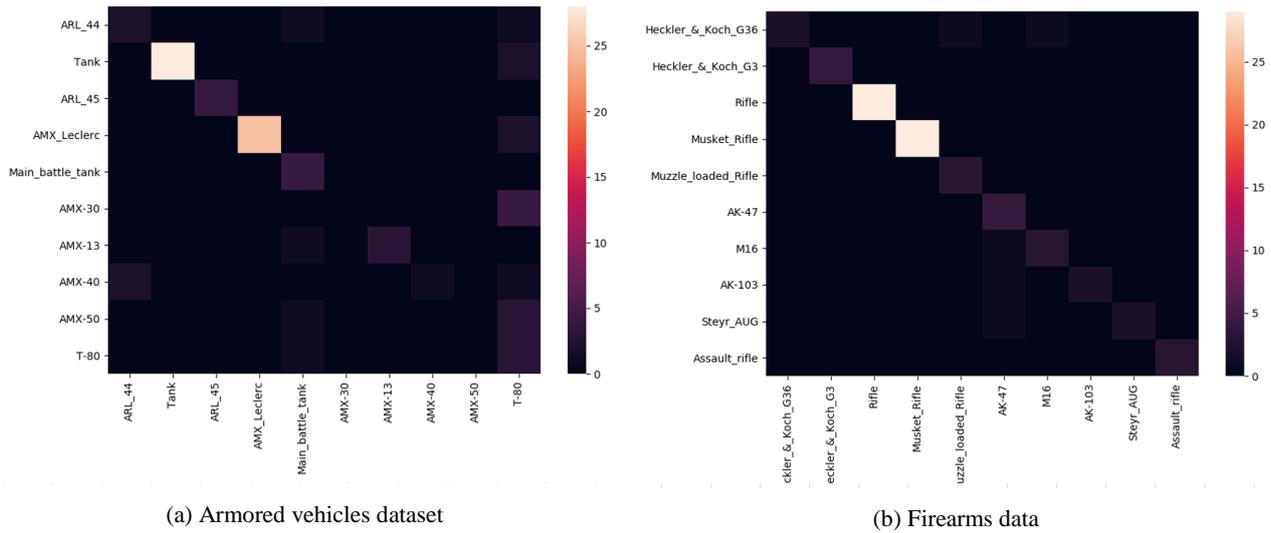

(a) Armored vehicles dataset   (b) Firearms data

Figure 6. Sample confusion matrices for 10 firearms and armored vehicles concepts.

### 3.3 Conclusions

Following the need of LEAs to be able to quickly adapt image classification systems to new concepts, we proposed in this section an extensible image classifications tool that relies on powerful convolutional neural networks and Web images. In order to illustrate the effectiveness of our approach on fine-grained classification tasks, we first automatically collected images for 314 firearms and 313 armored vehicles classes. Then, binary SVM classifiers were trained for each concept. We noted that, even though we used noisy Web images and few training examples, we were able to obtain accurate classifiers for the majority of concepts.

## 4. PERSON ATTRIBUTES

Sitting at the junction between face recognition and concept detection, person attributes detection usually involves specifically tailored methods for tasks, such as face landmark recognition or emotion detection. Another way to categorize attribute recognition methods is into two groups: global and local methods. Global methods extract features from the entire object, where accurate locations of object parts or landmarks are not required [30]. Local models [22][9] first detect object parts and extract features from each part. These local features are then concatenated to train classifiers. For instance, [22] predicted face attributes by extracting hand-crafted features from ten face parts, while Zhang et al. [12] recognized human attributes by employing a large number of pose related features [9]. Liu et al. [24] propose a novel deep learning framework, which combines massive objects and massive identities to pre-train two CNNs for face localization and attribute prediction, respectively. Here, we investigate the usefulness of generic CNN global image descriptors, as well as generic models in the case of person attributes.

### 4.1 Method

Most recent works concerning person features identification focus on dedicated deep learning architectures for face images. Given the scarcity of available training data and the need to quickly adapt to new features, we propose to treat the person features identification problem as a specific sub-task of concept classification. As in Section 3, we rely on the pipeline proposed in building semantic descriptors [14]. Here, we first evaluate a subset of Semfeat concepts, introduced in the previous section, that depict person features or objects that could serve as person identification (e.g. jewelry, clothing). Then, we investigate the impact of training dedicated person features classifiers. Both approaches are put to test on the CelebA [24], a large-scale person attributes image dataset.

The CelebA dataset offers more than 200K celebrity images, each with 40 attribute annotations. In order to assure a fair comparison of the two proposed set of classifiers, we select 7 attributes for which we find a match in ImageNet: eyeglasses, goatee, mustache, earrings, hat, necklace, necktie. In order to be able to capture more diverse styles of beards, we took the liberty to match goatee with the concept beard from ImageNet.

We propose two approaches for generating adapted concept classifiers. The first one relies on selecting relevant classifiers from the large set of ImageNet concepts without no further intervention. In order to expand the list of relevant ImageNet concepts, for each target concepts, we also look at the complete set of hyponyms, including intermediary and leaf nodes. This process expands the list of ImageNet concepts that are of interest from 7 to 94. At testing time, the prediction score for a target concept is given by selecting the maximum prediction score from its set of hyponyms.

ImageNet concepts are trained on a diverse set of images and may contain views that are detrimental for the task of person feature identification. For instance, the ImageNet images under the earrings concept are more likely to depict earrings in the foreground or in a store than a person wearing earrings, the situation that is of more interest for us. To counter this bias, we also train from scratch classifiers using only the training images from the CelebA dataset. There, all images present only people and are centered on people's faces. For comparability purposes, we use the same image descriptor as for the ImageNet concept classifiers, namely the fc7 layer of the VGG network.

### 4.2 Experiments and results

We first evaluate our person feature classifiers on the test set of the CelebA dataset. We look at the true positives rate (TP), true negatives rate (TN), false positives rate (FP), false negatives rate (FN), and the global accuracy (Acc.).

Table 2: Results of ImageNet models.

|      | Eyeglasses | Goatee | Moustache | Wearing Earrings | Wearing Hat | Wearing Neck-lace | Wearing Necktie | **Average** |
|------|------------|--------|-----------|------------------|-------------|-------------------|-----------------|-------------|
| **TP**   | 0.458 | 0.754 | 0.598 | 0.287 | 0.808 | 0.290 | 0.638 | **0.547** |
| **TN**   | 0.969 | 0.794 | 0.696 | 0.965 | 0.884 | 0.997 | 0.975 | **0.897** |
| **FP**   | 0.030 | 0.205 | 0.303 | 0.034 | 0.115 | 0.002 | 0.024 | **0.102** |
| **FN**   | 0.541 | 0.245 | 0.401 | 0.712 | 0.191 | 0.709 | 0.361 | **0.452** |
| **Acc.** | 0.714 | 0.774 | 0.647 | 0.626 | 0.846 | 0.643 | 0.806 | **0.722** |

In Table 2, we present the classification results of the ImageNet models. We notice that the generic classifiers perform adequately in terms of accuracy for most attributes. There is however a large gap between the rate of true positives and that of true negatives, confirming out observation that objects in ImageNet are portrayed mostly outside the context of people using them. The good performance is obtained by hat, as opposed to necklace or earring can be justified by their difference in size. A general model will more likely recognize a concept if it is clearly depicted in the image.

Table 3: Results of CelebA models.

|      | Eyeglasses | Goatee | Moustache | Wearing Earrings | Wearing Hat | Wearing Neck-lace | Wearing Necktie | **Average** |
|------|------------|--------|-----------|------------------|-------------|-------------------|-----------------|-------------|
| **TP**   | 0.782 | 0.758 | 0.584 | 0.844 | 0.933 | 0.767 | 0.888 | **0.794** |
| **TN**   | 0.921 | 0.866 | 0.890 | 0.622 | 0.979 | 0.663 | 0.923 | **0.837** |
| **FP**   | 0.078 | 0.133 | 0.110 | 0.377 | 0.020 | 0.336 | 0.076 | **0.162** |
| **FN**   | 0.217 | 0.241 | 0.415 | 0.155 | 0.066 | 0.232 | 0.111 | **0.205** |
| **Acc.** | 0.851 | 0.812 | 0.737 | 0.733 | 0.956 | 0.715 | 0.905 | **0.816** |

In Table 3, we show the classification results of the models trained using the images from the CelebA training set. As expected, the mean accuracy is higher than that obtained by the ImageNet classifiers but, surprisingly, the difference in not strikingly large. This result reinforces the obvious, that using task specific data is helpful but also suggests that generic classifiers can serve as good substitutes when lacking specific data.

Figure 7 illustrates prediction scores for the seven selected person identifiers provided by CelebA and ImageNet classifiers on two sample images. Although the confidence of the predictions may vary, we notice that both sets of classifiers manage to give positive predictions (score > 0.5) for clearly depicted attributes in the majority of cases.

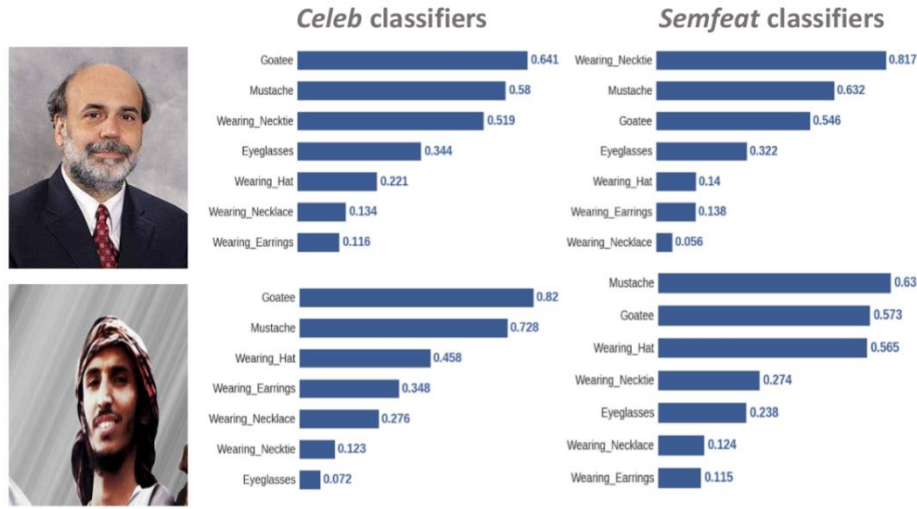

Figure 7. Prediction examples for the CelebA and Semfeat classifiers.

**4.3 Conclusions**

In this section, we proposed an alternative to the approaches for person attribute identification that use dedicated CNN architectures for this task. Circumventing the need for difficult obtainable annotated data and computationally expensive training procedures, we showed the capabilities of generic concept detectors on a subset of person attribute classes. We found that the generic models achieve similar performance to models trained on task specific data, with the evaluation being carried out on a well-known dataset.

## 5. INTERACTIVE ANNOTATION TOOL

An image annotation tool is important to define novel concepts and train concept detectors. The most efficient concept detectors are deep-learning based Convolutional Neural Networks (CNNs). For LEAs it is valuable to apply CNNs to find relevant concepts in the large collection of image and video data. A trained CNN can automatically detect a concept in an image and it generates a bounding box and a concept label for each detection. The concept label allows a query-by-keyword. The user can enter a keyword, which applies a filter on the concept label in a database, and analyze only the corresponding detections. The CNNs enable fast retrieval of relevant concepts in a large database. However, CNNs require many training examples to recognize a concept and it is very time consuming to annotate such images. Annotation of training examples is the process of indicating bounding boxes in images of that concept. The common approach is that a user manually indicates bounding boxes until there are sufficient training examples. The CNN is trained and when the performance is not sufficient yet, the user continues indicating more boxes manually. In this paper, we investigate another approach, where the user first draws new boxes manually and trains the detector. Then the detector is applied to more training images and the user validates – and optionally modifies – the automatically generated boxes and draws only missing boxes. In this section, we present a baseline approach of a random selection of images and the effects when only an extremely low number of training examples are provided.

**5.1 Method**

We developed an annotation tool where the user can annotate concepts and train a deep neural network to detect and localize these concepts in an image. The user can upload images with a graphical user interface (GUI) to annotate or detect concepts (Figure 8).

The annotation tool can be used in two modes:

- **Manual mode**: shows the image without any automatically generated bounding boxes. This is possible at all times.

- **Automatic mode**: shows the image with automatically generated bounding boxes. This is only possible when the tool contains a trained model.

The user can interact with the images and detections in several ways:

- **Accept**: The user can accept automatically generated bounding boxes.
- **Delete**: The user can delete automatically generated bounding boxes.
- **Modify**: The user can modify (move or resize) the automatically generated bounding boxes (when the box is misaligned).
- **Draw**: The user can manually draw a new bounding box in the image with a mouse (in manual mode or when the automatic system missed a detection).

The annotation tool currently uses the Single Shot multi-box Detector (SSD) network [25][6]. The SSD network generates detections. Each detection consists of a label (e.g., 'vehicle'), a bounding box to indicate the location in the image, and a confidence value between 0.0 and 1.0. This confidence can be used to threshold the resulting detections and remove false positives. The slider controls the threshold of which detections are visible.

After annotating a number of images, a neural network can be trained to detect the annotated concepts. All images that have a detection are used as an input to train the network. The first 3 layers are frozen [21] to avoid overtraining and to minimize the number of required annotations. The system trains for 20 epochs and stores the weights for each epoch. The weight file with the smallest loss is chosen as the weights to detect the concepts. Each time the network is trained the weights are reset to the pretrained weights. After the network is trained the tool can run the network on an image and show the concepts the network detected.

Initially, the network is trained on a small amount of images. The user can interact with the tool by accepting, deleting, modifying or drawing. This interaction leads to new confirmed bounding boxes that can be used to extend the train set. Retraining the network leads to improved performance of the model. This process can be repeated multiple times.

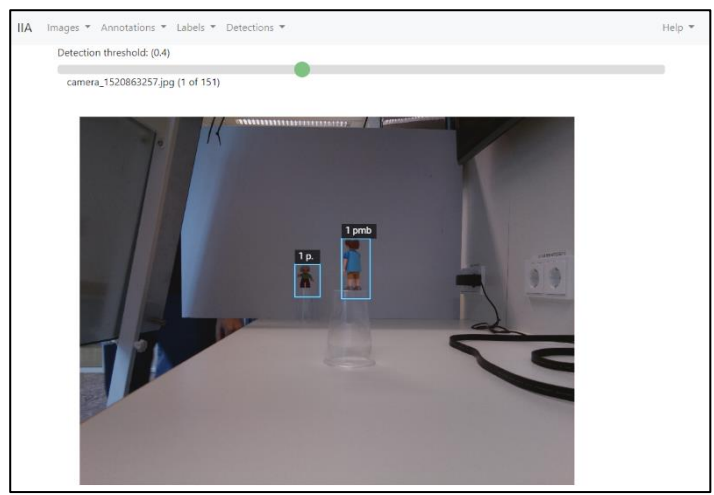

Figure 8. Overview of the GUI of the annotation tool.

### 5.2 Experimental setup

In the experiments we both validate the annotation tool and investigate whether re-training after a few annotations speeds up the annotation process without losing performance. Because our network is already trained on some commonly known image datasets, we use the FlickrLogo-32 dataset [31]. The network was originally not trained on logos, but should be feasible to transfer the network to the logo dataset after re-training on a few annotated images [6]. In the experiment, we ask the participants to annotate 50 images of each of the following 8 concepts: Adidas, Apple, BMW, Ferrari, Nvidia, Pepsi, Shell and Texaco (numbered concept 1 to 8 in this alphabetical order).

Four participants voluntarily participated in the experiment. The participants were not familiar with the annotation tool, but they received written instructions and a dummy concept to annotate to get familiar with the tool. Each of the participants was presented with two conditions: one with and one without the ability to re-train the system. In the condition without re-training, the participants manually annotated 50 images for one concept. In the condition with re-training, the participants manually annotated 20 images, re-trained the system, and annotated the remaining 30 images in the automatic mode. These images did already contain suggested bounding boxes. To adjust for learning rate and fatigue of the participants, the 50 images were presented in random order. Half of the participants started with the first condition and the other half of the participants started with the second condition. The concepts were ordered using Latin Square [12] to make the experiment balanced (Table 4). After each concept the participants were allow to take a break.

Table 4: Latin Square for 8 concepts (1 – 8) and four volunteers (P1 – P4) with manual mode (white fill) and automatic mode (gray fill).

|    | Concept order | | | |
|----|------|------|------|------|
| **P1** | 1, 2 | 3, 4 | 5, 6 | 7, 8 |
| **P2** | 3, 4 | 7, 8 | 1, 2 | 5, 6 |
| **P3** | 8, 7 | 6, 5 | 4, 3 | 2, 1 |
| **P4** | 6, 5 | 2, 1 | 8, 7 | 4, 3 |

We logged the drawn bounding boxes, set a timer and logged the events (going to a new image, start and end of training, annotated boxes), and asked the participants to fill in an evaluation to capture the usability of the system. User satisfaction is measured using the System Usability Scale (SUS) [10].

### 5.3 Results

The images appear to contain on average 1.3 bounding boxes per image.

First, we observe the learning effect. The results are shown in Figure 9. This figure shows the total annotation time on the vertical axis and the annotated concept from first to last on the horizontal axis. The total annotation time also includes the training time of 89 seconds.

Each volunteer annotated 8 concepts. The first concept took 46% more time than the average time for the other concepts. The first concept of the automatic mode (overall the first or the fifth concept) was 34% slower than the other concepts in the same mode and the first concept of the manual mode (again the first or the fifth concept) was 18% slower than other concepts in the same mode. In the manual mode, we do not observe a continuation in the learning effect after the first concept, however, in the automatic mode, we do observe a continuation of the learning effect. The Latin square was used to avoid the influence of learning effects in the comparison between manual mode and automatic mode.

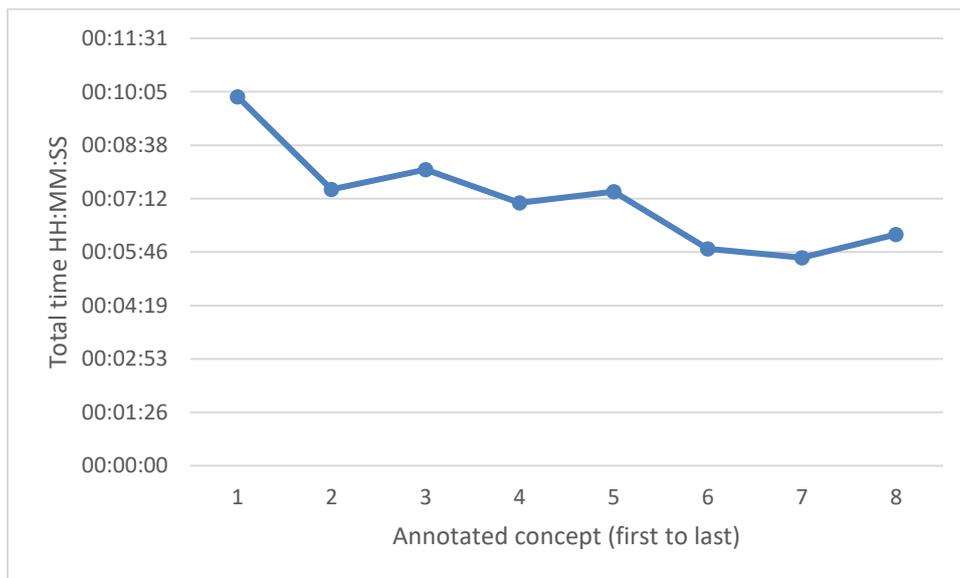

Figure 9. Learning curve for all four participants.

In order to compare the manual mode with the automatic mode, the annotation times are measured. For manual mode, the time is measured for all images, and for automatic mode, the time is measured for 20 images before training and for 30 images after training. The results are shown in Table 5. The table shows that manual mode requires 4 seconds per detection and that this is 2.6 times faster than the automatic mode. Most surprising is that the 20 images before training are also slower than the manual mode. The average time per detection before and after training increases from 7 to 10 seconds (factor 1.4). In the discussion (Sec. 5.5) we come back on these observations.

Table 5: Total times (M:SS) for each person in manual and automatic mode.

| Mode | Manual | Automatic | |
|---|---|---|---|
| # images | 50 images | 20 images before training | 30 images after training |
| **P1** | 3:28 | 1:37 | 3:52 |
| **P2** | 4:59 | 2:56 | 6:41 |
| **P3** | 3:48 | 4:25 | 6:57 |
| **P4** | 3:41 | 2:38 | 7:32 |
| **Average total time (M:SS)** | 3:59 | 2:54 | 6:16 |
| **Average time per detection (sec)** | 3.6 sec | 6.7 sec | 9.6 sec |

User evaluation is measured using the System Usability Scale (SUS) [10]. The participants evaluated the system in both conditions (manual and automatic mode). The results show a usability score of 72 for the system with re-training (min = 65, max = 80), and 92 for the system without re-training (min = 90, max = 95). According to SUS, a score above 68 is considered above average [10]. This result shows that the system without re-training is preferred over the system with re-training in terms of usability. In the subjective evaluation, the main reason is the increased complexity of the tool with the re-training option.

### 5.4 Recent progress with active learning

The baseline approach to select new images uses random image selection. Recently, two active learning approaches were compared to the baseline approach [7]. The first uses the high-confidence sampling [20] to allow rapid validation and minimize drawing and modification of bounding boxes. The second uses uncertain detections close to the decision boundary [35][11] to select the most informative samples. The results are summarized in Table 6. The results show that High-confidence sampling reaches the highest mAP value (23% higher than random baseline) in minimal time (46% less time than random baseline). Both active-learning approaches perform better than the random baseline approach.

Table 6: Results from active learning in the annotation tool after 2000 training samples.

| | mAP (%) at 250 images | Time (minutes) at mAP=20% |
|---|---|---|
| **Baseline: Random sampling** | 17 | 13 |
| **Active: Uncertainty sampling** | 16 | 16 |
| **Active: High-confidence sampling** | **21** | **7** |

### 5.5 Discussion

We observed that in the original Latin square (Table 4) concepts {5,6} are present twice in the automatic mode and missing in the manual mode and for concepts {3,4} it is vice versa. Therefore, we also analyzed the results based on a subset of only the concepts {1,2,7,8}. These concepts are well balanced over the manual and the automatic mode. The results are shown in Table 7. This slightly changes the conclusions from earlier results. The average time per detection in the automatic mode (8 seconds) is 2 times slower than the manual mode (4 seconds). The average time per detection before and after training increases from 7 to 8 seconds (factor 1.2). The Latin square can be improved with another assignment of manual mode and automatic mode (Table 8).

Table 7: Total times (M:SS) for each person in manual and automatic mode for original Latin square using only concepts {1,2,7,8}.

| Mode | Manual | Automatic | |
|---|---|---|---|
| # images | 50 images | 20 images before training | 30 images after training |
| **Average total time (M:SS)** | 4:31 | 2:55 | 5:19 |
| **Average time per detection (sec)** | 4.2 sec | 6.7 sec | 8.2 sec |

Table 8: Improved Latin Square for 8 concepts (1 – 8) and four volunteers (P1 – P4) with manual mode (white fill) and automatic mode (gray fill).

|  | Concept order | | | |
|---|---|---|---|---|
| **P1** | 1, 2 | 3, 4 | 5, 6 | 7, 8 |
| **P2** | 3, 4 | 7, 8 | 1, 2 | 5, 6 |
| **P3** | 8, 7 | 6, 5 | 4, 3 | 2, 1 |
| **P4** | 6, 5 | 2, 1 | 8, 7 | 4, 3 |

In the previous tables and figures, the average time per detection is computed by dividing the total time by the number of detections. Therefore, the time per detection includes all other activities, such as modifying the threshold and applying the concept detector. The time for a detection can also be computed by using only the time between the moment that a detection was added and the moment of another previous activity (e.g., loading an image, running the detector or adding a previous detection). The time per detection that excludes other activities are shown in Table 9. The table shows that the estimated times per detection in manual mode is not faster than in automatic mode. So, adding a detection is not slower in the mode after training but it is related to other activities. One of the other activities is labelled as 'loading of images'. The image loading is activated when going to the next image or when applying the detector (e.g., after modifying the threshold). These results are also shown in Table 9. The table shows that the users have increased image loading activity after training, it increases from 1.1 to 2.4. To obtain an efficient annotation tool, the other interaction activities must be minimized. In the mode after training, 17% is generated automatically and 83% of the detections is generated manually. One of the crucial parts to obtain significant improved performance with the annotation tool is a good detector that automatically generates many acceptable detections.

Table 9: Average time per detection (sec) in manual and automatic mode excluding other activity.

| Mode | Manual | Automatic | |
|---|---|---|---|
| # images | 50 images | 20 images before training | 30 images after training |
| **Average time per detection (sec)** | 3.4 sec | 2.3 sec | 2.2 sec |
| **Average number of image loads per image (incl. run detector)** | 1.1 | 1.1 | 2.4 |
| **Percentage automatic detections** | 0% | 0% | 17% |

It is naive to assume that the user always benefits from automatically generated bounding boxes. The experiments show that the manual mode is faster than the automatic mode. The user could benefit from the system when automatically generated bounding boxes only need to be validated (accept or delete). Modification of the bounding boxes is time consuming. So, in the automatic mode, missing boxes are not the largest problem (performance is equal to manual mode) but false positives or misaligned detections need to be avoided (because this is additional burden for the user in comparison to manual mode). A future experiment should focus on active learning strategies to learn faster than random selection and continue until higher mAP values to improve correct detections and a high threshold to avoid false detections.

Future experiments should also include a proper Latin Square set-up with more detailed measurements and logging to analyze why the automatic mode is faster or slower than manual mode. This requires counting how many detections are accepted, detected, drawn or modified, how often the threshold is modified or the detector is applied, and how much time each of these interaction costs. In the experimental setup we only measured the number of detections and not the interaction type for each detection. Furthermore, the interaction must be simple to accept a few detections and remove the others, or vice versa remove a few and accept the rest. We used 3 frozen layers in the SSD network. In cases with only very few annotations, it may be necessary to freeze more layers and in cases with many annotations, it may be allowed to freeze less layers.

### 5.6 Conclusions

In this section, we presented an image annotation tool. The tool is important to allow LEAs to flexibly adapt to novel concepts that are relevant in their cases. We show that manual annotation can be more efficient than annotation assisted with automatically generated detections on randomly selected images. Deletion of false detections and modification of misaligned detections can be time consuming. Recent progress showed that active learning approaches can reach better results than random image selection.

## 6. QUERY EXPANSION

Concepts detections, like SSD in the previous section, allow query-by-keyword. A user enters a keyword and only images that contain detections with the corresponding concept label are retrieved. This allows users to quickly find a concept in a large database with images [4]. However, the user needs to know which keywords are valid and spelling mistakes are not allowed. Query expansion expands the keyword to similar words, such as synonyms (other words with the same meaning), hypernyms and hyponyms (parents and children of the concept). Query expansion is relevant because it allows a user to enter a keyword without knowing the exact labels of the concept detectors. The expansion finds similar words that can be matched to the concept labels. This makes a concept detection tool more flexible and easier to use.

### 6.1 Method

We use an automatic query-to-concept mapping to find relevant concepts such as synonyms. This method can either be used to map the query to the pre-trained concepts that are available in a Concept Bank, or to map the query to the nearest neighbors (assuming that all words are present as concepts). We provide a system that uses two tools to create the mapping.

The first tool is WordNet[*], which is a large English lexicon, and falls into the ontology category. In our implementation we map the first sense of the query, if the query exists in WordNet, to the first sense of each concept label. We calculate the Wu-Palmer similarity (WUP) to determine the weight for each label. The WUP similarity uses the depth of the two senses in the tree and the depth of their Least Common Subsumer (the most specific parent node that connects the two senses) to determine the similarity. We use the implementation of the wordnet package within NLTK[†] in Python 3.5 for the calculation of the similarity. This similarity does, therefore, not only output synonyms, but also hypernyms and hyponyms (parents and children of the concept) and other related concepts. The output of the module is the top k concepts that best match (have the highest similarity to) the query.

Although WordNet is a good resource for query-to-concept mapping, it is not complete. We, therefore, added a second tool based on a machine-learning technique. We use word2vec, which produces semantic embeddings. The models of Word2vec either use skip-grams or continuous bag of words (CBOW) to create neural words embeddings using a shallow neural network that is trained on a huge dataset, such as GoogleNews, Twitter, Wikipedia or Gigawords. Each word vector is trained to maximize the log probability of neighboring words. This allows for a good performance in associations, such as king – man + woman = queen. These word vectors, thus, allow us to relate words, but the specific relation between the words is not explicit using this model. In our implementation we use the Gensim code[‡], which is an implementation of the skip-gram models of Mikolov et al. [27], following the baseline implementation of de Boer et al. [5]. As a background model we use the pre-trained model on GoogleNews, in which words are represented as a 300 dimensional vector. We calculate the cosine similarity between the user query and each of the concept labels, and output the top k concepts that best match (have the highest similarity to) the query.

### 6.2 Experiments and results

In our experiments, we use the 1000 ImageNet concepts [13] as our Concept Bank. We test several related concepts. In WordNet the query "gun" provides the following top three concepts as most similar: "projectile", "revolver" and "assault rifle". The query "car" relates to "limousine", "beach wagon" and "sports car". In word2vec, the query "gun" is mapped to "assault rifle", "rifle" and "revolver" and the query "car" is mapped to "car wheel", "passenger car" and "tow car". The results are shown in Table 10. These results show that both methods are able to find related words. The performance of the methods is dependent on the words available in the Concept Bank, i.e. in case of searching for a synonym, this synonym should be present in the Concept Bank.

---

[*] https://wordnet.princeton.edu/

[†] http://www.nltk.org/_modules/nltk/corpus/reader/wordnet.html

[‡] https://radimrehurek.com/gensim

Table 10: Example query keywords and related concepts.

| Method | Query keyword | Related concepts |
|---|---|---|
| **WordNet** | Gun | "projectile", "revolver", "assault rifle", … |
|  | Car | "limousine", "beach wagon", "sports car", … |
| **Word2Vec** | Gun | "assault rifle", "rifle", "revolver", … |
|  | Car | "car wheel", "passenger car", "tow car", … |

### 6.3 Conclusions

The semantic analysis presented in this section is relevant because there is a mismatch between the concepts that people use to formulate their query and the concepts that are used to index the image or video with. We showed that it is possible to map a user query to concepts that are known to the system or vice versa.

## 7. CONCLUSIONS

In this paper, we showed five contributions. The first contribution allows image-based GPS localization to estimate the origin of an image. The second contribution enables analysis of fine-grained concepts to distinguish sub-categories in a generic concept. The third contribution is the recognition of person attributes (e.g., glasses or moustache) to enable query by textual description for a person. The fourth contribution is an image annotation tool based with a manual and an automatic mode. The fifth contribution increases the flexibility for LEAs in the query definition by using query expansion.

## ACKNOWLEDGEMENT

The ASGARD ("Analysis System for Gathered Raw Data") project has received funding from the EU H2020 Secure Societies program (FCT-01-2015) under grant agreement number 700381. Any opinions expressed in this paper do not necessarily reflect the views of the European Community. The Community is not liable for any use that may be made of the information contained herein. In this joint publication, CEA contributed sections 2, 3 and 4 and TNO contributed Sections 5 and 6.